# Towards Normalizing the Edit Distance Using a Genetic Algorithms–Based Scheme

Muhammad Marwan Muhammad Fuad

Department of Electronics and Telecommunications
Norwegian University of Science and Technology (NTNU)
NO-7491 Trondheim, Norway
`marwan.fuad@iet.ntnu.no`

**Abstract:** The normalized edit distance is one of the distances derived from the edit distance. It is useful in some applications because it takes into account the lengths of the two strings compared. The normalized edit distance is not defined in terms of edit operations but rather in terms of the edit path. In this paper we propose a new derivative of the edit distance that also takes into consideration the lengths of the two strings, but the new distance is related directly to the edit distance. The particularity of the new distance is that it uses the genetic algorithms to set the values of the parameters it uses. We conduct experiments to test the new distance and we obtain promising results.

## 1  Introduction

*Similarity search* is an important problem in computer science. This problem has many applications in data mining, computational biology, pattern recognition, and others. In this problem a pattern or a *query* is given and the task is to retrieve the data objects in the database that are "close" to that query according to some semantics that quantify that closeness. This closeness or similarity is depicted using a principal concept which is the *similarity measure* or its more powerful form; the *distance metric*.

Because of its topological properties, the metric model (reflexivity, non-negativity, symmetry, triangle inequality) has been widely used to process similarity queries, but later other models were proposed.

The *edit distance* is the main distance used to measure the similarity between two strings. It is defined as the minimum number of delete, insert, and change operations needed to transform string *S* into string *T*.

But the edit distance has its limitations because it considers local similarity only and does not imply any global level of similarity.

This work was carried out during the tenure of an ERCIM "Alain Bensoussan" Fellowship Programme. This Programme is supported by the Marie-Curie Co-funding of Regional, National and International Programmes (COFUND) of the European Commission.

In [10], [11], and [12] we presented two new extensions of the edit distance. These new extensions consider a global level of similarity which the edit distance didn't consider. But the parameters used with these two distances were defined using basic heuristics, which substantially limited the search space.

In this paper we propose a new extension of the edit distance. This extension aims to normalize the edit distance using an approach that relates it directly to the edit distance. The new distance uses the genetic algorithms as an optimization method to set the parameters it uses.

Section 2 of this paper presents the related work. Section 3 introduces the new distance. Section 4 validates it through different experiments and Section 5 concludes the paper.

## 2  Related Work

*Strings*, also called *sequences* or *words*, are a way of representing data. This data type exists in many fields of computer science such as molecular biology where DNA sequences are represented using four nucleotides which correspond to the four bases: adenine (A), cytosine (C), guanine (G) and thymine (T). This can be expressed as a 4-symbol alphabet. Protein sequences can also be represented using a 20-symbol alphabet which corresponds to the 20 amino acids.

Written languages are also expressed in terms of alphabets with letters (26 in English). Spoken languages are represented using phonemes (40 in English). Texts use alphabets with a very large size (the vocabulary items of a language). These examples show that strings are ubiquitous.

One of the main distances used to handle sequential data is the edit distance (ED) [13], also called the *Levenshtein distance*, which is defined as the minimum number of delete, insert, and substitute operations needed to transform string $S$ into string $R$.

Formally, ED is defined as follows: Let $\Sigma$ be a finite alphabet, and let $\Sigma^*$ be the set of strings on $\Sigma$. Given two strings $S = s_1 s_2 .... s_n$ and $R = r_1 r_2 .... r_m$ defined on $\Sigma^*$. An *elementary edit operation* is defined as a pair: $(a,b) \neq (\lambda, \lambda)$, where $a$ and $b$ are strings of lengths 0 and 1, respectively. The elementary edit operation is usually denoted $a \rightarrow b$ and the three elementary edit operations are $a \rightarrow \lambda$ (deletion) $\lambda \rightarrow b$ (insertion) and $a \rightarrow b$ (substitution). Those three operations can be weighted by a weighting function $\gamma$ which assigns a nonnegative value to each of these operations. This function can be extended to edit transformations $T = T_1 T_2 ... T_m$.

The edit distance between $S$ and $R$ can then be defined as:

$$ED(S, R) = \{\gamma(T) | T \text{ is an edit transformation of } S \text{ into } R\} \qquad (1)$$

ED is the main distance measure used to compare two strings and it is widely used in many applications. Fig. 1 shows the edit distance between the two strings $S_1 = \{M, A, R, W, A, N\}$ and $S_2 = \{F, U, A, D\}$

|   |   | M | A | R | W | A | N |
|---|---|---|---|---|---|---|---|
|   | 0 | 1 | 2 | 3 | 4 | 5 | 6 |
| F | 1 | 1 | 2 | 3 | 4 | 5 | 6 |
| U | 2 | 2 | 2 | 3 | 4 | 5 | 6 |
| A | 3 | 3 | 2 | 3 | 4 | 4 | 5 |
| D | 4 | 4 | 3 | 3 | 4 | 5 | **5** |

**Fig. 1.** The edit distance between two strings.

ED has a few drawbacks; the first is that it is a measure of local similarities in which matches between substrings are highly dependent on their positions in the strings [5]. In fact, the edit distance is based on local procedures both in the way it is defined and also in the algorithms used to compute it. Another drawback is that ED does not consider the length of the two strings.

Several modifications have been proposed to improve ED. In [10], [11], and [12] two new extensions of ED; the *extended edit distance* (EED) and the *multi-resolution extended edit distance* (MREED) were proposed. These two distances add a global level of similarity to that of ED by including the frequency of characters or bi-grams when computing the distance. The problem with these two distances is that they use parameters which are set using basic heuristics which makes the search process ineffective.

It is worth mentioning that the two distances EED and MREED, as well as ED, are all metric distances.

Another important modification is the *normalized edit distance* (NED) [8]. The rationale behind this distance is that the length of the two strings should be taken into account when computing the distance between them. An *editing path P* between two strings $S$ and $R$, of lengths $n$ and $m$, respectively ($n \leq m$), is a sequence of ordered pairs of integers $(i_k, j_k)$, where $0 \leq k \leq m$, that satisfies the following:

i-  $0 \leq i_k \leq |S|$, $0 \leq j_k \leq |R|$;
    $(i_0, j_0) = (0,0), (i_m, j_m) = (|S|, |R|)$

ii- $0 \leq i_k - i_{k-1} \leq 1$, $0 \leq j_k - j_{k-1} \leq 1$, $\forall k \geq 1$

iii- $i_k - i_{k-1} + j_k - j_{k-1} \geq 1$

The weights can be associated to paths as follows:

$$\omega(P_{S,R}) = \sum_{k=1}^{m} \gamma(S_{i_{k-1}+1...i_k} \to R_{j_{k-1}+1...j_k})$$

It follows that:

$$ED(S, R) = \min\{\omega(P) \mid P \text{ is an edit transformation of } S \text{ into } R\}$$

Let $\hat{\omega}(P) = \omega(P)/L(P)$, where $L$ is the length of $P$, the normalized edit distance NDE is defined as:

$$NDE(S, R) = \min\{\hat{\omega}(P)\} \qquad (2)$$

An important notice about this definition is that it is expressed in terms of paths and not in terms of the edit operations.

It has been shown in [8] that NDE is not a distance metric.

## 3 Genetic Algorithms-based Normalization of the Edit Distance (GANED)

ED we presented in Section 2 was mainly introduced to apply to spelling errors. This makes the edit operations a main component of ED. NDE, although takes into consideration the lengths of the two strings, which is an important modification in our opinion, is based on a different principle than that of ED, which, we think, causes it to lose some of the principal characteristics of ED.

In this work we present a new modification of the edit distance that also takes the lengths of the strings into account. However, our proposed distance uses a completely different approach than that of NDE. Our new distance is directly related to the edit distance. In fact, the new distance is a lower bound of the edit distance.

### 3.1 GANED

Let $\Sigma$ be a finite alphabet, and let $\Sigma^*$ be the set of strings on $\Sigma$. Let $n$ be an integer, and let $f_{a_n}^{(S)}$ be the frequency of the $n$-gram $a_n$ in $S$, and $f_{a_n}^{(T)}$ be the frequency of the $n$-gram $a_n$ in $T$, where $S$, $T$ are two strings in $\Sigma^*$.

The GANED distance between $S$ and $T$ is defined as:

$$GANED(S,T) = ED(S,T) \times \left[1 - \frac{2\sum_{n=1}^{\max(|S|,|T|)} \lambda_n \left(\sum_{a_n \in A^n} \min\left(f_{a_n}^{(S)}, f_{a_n}^{(T)}\right) + (n-1)\right)}{n(|S|+|T|)}\right] \qquad (3)$$

where $|S|, |T|$ are the lengths of the two strings $S, T$ respectively, and where $\lambda_n \in [0,1]$. $\lambda_n$ are called the frequency factors.

Notice that

$$0 \leq 2 \sum_{n=1}^{max(|S|,|T|)} \lambda_n \left( \sum_{a_n \in A^n} min\left(f_{a_n}^{(S)}, f_{a_n}^{(T)}\right) + (n-1) \right) \leq n(|S|+|T|)$$

So ED is multiplied by a factor whose value varies between 0 and 1, so GANED as presented in (3) includes a form of normalization. In fact: $0 \leq GANED(S,R) \leq ED(S,R)$ can be written as $0 \leq \frac{GANED(S,R)}{ED(S,R)} \leq 1$ which is the common form of normalization, and we could have expressed our new distance in the latter form. However, instead of imposing a condition that the two strings be different (thus ED=0), we preferred to introduce the new distance in the form shown in (3).

Notice also that GANED is a lower bound of ED, so the relation between the two distances is direct.

As mentioned in Section 1, the idea of considering the frequency of characters or bi-grams in computing the distance has previously been proposed in [10], [11], and [12]. However, the definition of the frequency factors remains problematic in these three works. On the one hand, the search space they use is very limited, and on the other hand, generalizing the distances proposed in [10], [11], and [12] using the same basic heuristics to define the frequency factors makes this process inefficient yet limited to very small regions in the search space.

GANED uses one very powerful optimization method; the genetic algorithms, to define the frequency factors $\lambda_n$. The use of the genetic algorithms makes the search more effective.

### 3.2 The Genetic Algorithms

The *Genetic Algorithms* are a member of a large family of stochastic algorithms called *Evolutionary Algorithms* (EAs) which are population-based optimization algorithms inspired by nature, particularly the theory of evolution. In Fig. 2 we show the members of the EAs. These members differ in implementation but they use the same principle.

Of the EAs family, GAs are the most widely known. GAs have the following elements: a population of individuals (also called *chromosomes*), selection according to fitness, crossover to produce new offspring, and random mutation of new offspring [9]. GAs create an environment in which a population of individuals, representing solutions to a particular problem, is allowed to evolve under certain rules towards a state that minimizes, in terms of optimization, the value of a function which is usually called the *fitness function* or the *objective function*.

There are a large number of variations of GAs. In the following we present a description of the simple, classical GAs. GAs start by defining the problem variables

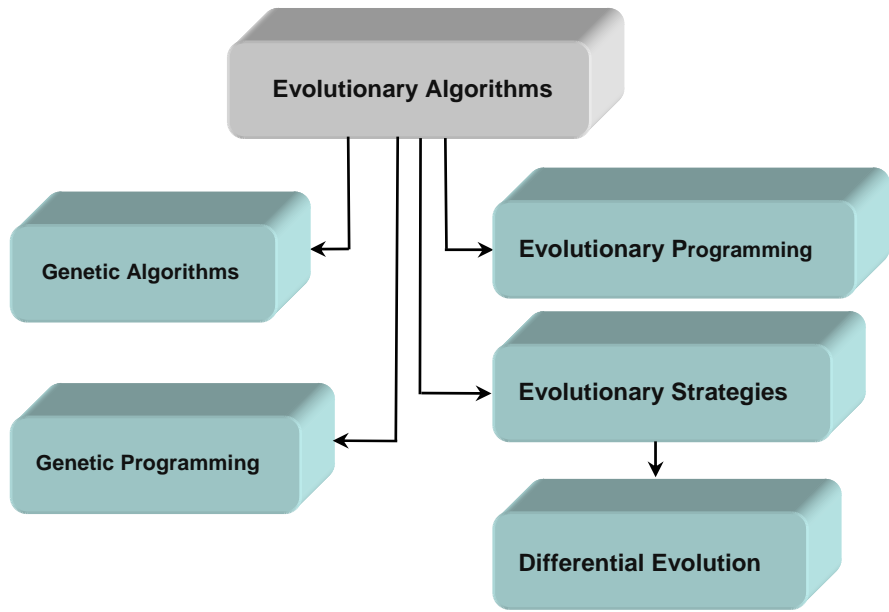

**Fig. 2.** The family of evolutionary algorithms

(*genes*) and the fitness function. These variables can be bounded or unbounded. A particular combination of variables produces a certain value of the fitness function and the objective of GAs is to find the combination that gives the best value of the fitness function. The terminology "best" implies that there is more than one solution and the solutions are not of equal value [2].

After defining the variables and the fitness function GAs start by randomly generating a number *pSize* of individuals, or *chromosomes*. This step is called *initialization*.

GAs were originally proposed to be binary coded to imitate the genetic encoding of natural organisms [15]. But later other encoding schemes were presented. The most widely used scheme is real-valued encoding. In this scheme a candidate solution is represented as a real-valued vector in which the dimension of the chromosomes is constant and equal to the dimension of the solution vectors [1]. This dimension is denoted by *nPar*. The fitness function of each chromosome is evaluated. The next step is *selection*. The purpose of this procedure is to determine which chromosomes are fit enough to survive and possibly produce offspring. This is decided according to the fitness function of the chromosome in that the higher the fitness function is the more chance it has to be selected for mating. There are several selection methods such as the *roulette wheel selection*, *random selection*, *rank selection*, *tournament selection*, and others [9]. The percentage of chromosomes selected for mating is denoted by *sRate*. *Crossover* is the next step in which offspring of two parents are produced to enrich the population with fitter chromosomes. There are several approaches to perform this process, the most common of which is *single-point* crossover and *multi-point* crossover.

While crossover is the mechanism that enables the GA to communicate and share information about fitter chromosomes, it is not sufficient to efficiently explore the search space. *Mutation*, which is a random alteration of a certain percentage *mRate* of chromosomes, is the other mechanism which enables the GA to examine unexplored regions in the search space. It is important to keep a balance between crossover and mutation. High crossover rate can cause converging to local minima and high mutation rate can cause very slow convergence.

Now that a new generation is formed, the fitting function of the offspring is calculated and the above procedures repeat for a number of generations *nGen* or until a stopping criterion terminates the algorithm.

## 4 Empirical Evaluation

We tested the new distance GANED on time series because this is our field of expertise, but we believe GANED is highly applicable in bioinformatics and text mining.

Time series data are normally numeric, but there are different methods to transform them to symbolic data. The most important symbolic representation method of time series is the *Symbolic Aggregate approXimation* (SAX) [7]. The first step of SAX is to normalize the time series because SAX is based on the assumption that normalized time series have a Guassian distribution, so SAX can only be applied to normalized time series. The second step is to reduce the dimensionality of the time series by using a time series representation method called *Piecewise Aggregate Approximation* (PAA) [3], [14]. This PAA representation of the time series is then discretized. This is achieved by determining the breakpoints. The number of the breakpoints is related to the desired alphabet size and their locations are obtained using statistical lookup tables.

The distance between the resulting time series after applying the above steps is computed using the following relation:

$$MINDIST(\hat{S}, \hat{R}) \equiv \sqrt{\frac{n}{N}} \sqrt{\sum_{i=1}^{N}(dist(\hat{s}_i, \hat{r}_i))^2} \qquad (4)$$

Where $n$ is the length of the original time series, $N$ is the number of segments, $\hat{S}$ and $\hat{R}$ are the symbolic representations of the two time series $S$ and $R$, respectively, and where the function $dist(\ )$ is implemented by using the appropriate lookup table.

We tested our new distance GANED on time series classification task based on the first nearest-neighbor (1-NN) rule using leaving-one-out cross validation. This means that every time series is compared to the other time series in the dataset. If the 1-NN does not belong to the same class, the error counter is incremented by 1.

We conducted experiments using datasets of different sizes and dimensions available at UCR [4].

As indicated earlier, we tested GANED on symbolically represented time series This means that the time series were transformed to symbolic sequences using the first three step of SAX presented earlier in this section, but instead of using MINDIST given in relation (4), we use our distance GANED. The parameters $\lambda_n$ in the definition of GANED (relation (3)) are defined using GAs. This means that for each value of the alphabet size we formulate a GAs optimization problem where the fitness function is the classification error, and the parameters of the optimization problem are $\lambda_n$. Practically $n$ can take any value that does not exceed that of the shortest string of the two strings $S$, $T$. However, in the experiments we conducted $n \in \{1,2,3\}$ because these are the values of interest for time series. The values of $\lambda_n$ varied in the interval $[0,1]$.

Notice that GANED can be applied to strings of different lengths, which is one of its advantages since most similarity measures in time series mining are applied only to time series of the same length.

For the GAs we used, the population size *pSize* was 12, the number of generations *nGen* was set to 20. The mutation rate *mRate* was 0.2 and the selection rate *sRate* was 0.5. The dimension of the problem *nPar* depends on the number of parameters used in GANED (as mentioned earlier, they were tested for $n \in \{1,2,3\}$). Table 1 summarizes the symbols used in the experiments together with their corresponding values.

For each dataset we use GAs on the training datasets to get the vector $\lambda_n$ that minimizes the classification error on these training datasets, and then we utilize this optimal $\lambda_n$ vector on the corresponding testing datasets to get the final classification error for each dataset.

We compared GANED with MINDIST. This means after we represent the time series symbolically as indicated at the beginning of this section we classify them using GANED first then using MINDIST. We chose to compare GANED with MINDIST because we used the same symbolic representation that MINDIST uses. However, GANED can be used with any sequential data.

It is important to mention however that MINDIST has a lower complexity than that of GANED.

In Table 2 we present some of the results we obtained for alphabet size equal to 3, 10, and 20, respectively.

**Table 1.** The symbol table together with the corresponding values used in the experiments.

| *pSize* | Population size | 12 |
|---|---|---|
| *nGen* | Number of generations | 20 |
| *mRate* | Mutation rate | 0.2 |
| *sRate* | Selection rate | 0.5 |
| *nPar* | Number of parameters | varies |

**Table 2.** Comparison between the classification error of GANED and MINDIST

**CBF**

| $\alpha^*$ | n-gram | GANED | | MINDIST |
|---|---|---|---|---|
| | | $\lambda_n$ | Classification Error | Classification Error |
| 3 | 1 | [0.77491] | 0.026 | 0.382 |
| | 2 | [0.81776   0.87965] | 0.026 | |
| | 3 | [0.93285   0.97274   0.75836] | 0.023 | |
| 10 | 1 | [0.43021] | 0.031 | 0.104 |
| | 2 | [0.34446   0.32247] | 0.031 | |
| | 3 | [0.13412   0.45606   0.080862] | 0.039 | |
| 20 | 1 | [0.37819] | 0.053 | 0.088 |
| | 2 | [0.8962   0.72086] | 0.079 | |
| | 3 | [0.96216   0.091513   0.83706] | 0.062 | |

$\alpha^*$:alphabet size

**Coffee**

| $\alpha$ | n-gram | GANED | | MINDIST |
|---|---|---|---|---|
| | | $\lambda_n$ | Classification Error | Classification Error |
| 3 | 1 | [0.81472] | 0.179 | 0.464 |
| | 2 | [0.43021   0.22175] | 0.214 | |
| | 3 | [0.21868   0.19203   0.34771] | 0.214 | |
| 10 | 1 | [0.89292] | 0.179 | 0.464 |
| | 2 | [0.97059   0.93399] | 0.214 | |
| | 3 | [0.4899   0.81815   0.08347] | 0.143 | |
| 20 | 1 | [0.12393] | 0.107 | 0.143 |
| | 2 | [0.16825   0.9138] | 0.107 | |
| | 3 | [0.81472   0.95717   0.67874] | 0.107 | |

**Face Four**

| $\alpha$ | n-gram | GANED | | MINDIST |
|---|---|---|---|---|
| | | $\lambda_n$ | Classification Error | Classification Error |
| 3 | 1 | [0.022414] | 0.057 | 0.239 |
| | 2 | [0.57462   0.57425] | 0.057 | |
| | 3 | [0.2038   0.60654   0.38334] | 0.057 | |
| 10 | 1 | [0.015908] | 0.045 | 0.182 |
| | 2 | [0.16625   0.34168] | 0.057 | |
| | 3 | [0.57997   0.3957   0.21003] | 0.057 | |
| 20 | 1 | [0.54483] | 0.114 | 0.193 |
| | 2 | [0.92995   0.18334] | 0.102 | |
| | 3 | [0.57758   0.28758   0.15406] | 0.090 | |

**Table 2 (continued)**

**Gun_Point**

| α | n-gram | GANED $\lambda_n$ | Classification Error | MINDIST Classification Error |
|---|---|---|---|---|
| 3 | 1 | [0.19728] | 0.193 | 0.307 |
| 3 | 2 | [0.95798  0.58518] | 0.193 | |
| 3 | 3 | [0.40628  0.95213  0.68035] | 0.2 | |
| 10 | 1 | [0.98445] | 0.147 | 0.233 |
| 10 | 2 | [0.93927  0.99038] | 0.127 | |
| 10 | 3 | [0.15187  0.40029  0.24364] | 0.12 | |
| 20 | 1 | [0.16625] | 0.06 | 0.12 |
| 20 | 2 | [0.5852  0.0038735] | 0.06 | |
| 20 | 3 | [0.32809  0.42736  0.12747]] | 0.06 | |

**Olive Oil**

| α | n-gram | GANED $\lambda_n$ | Classification Error | MINDIST Classification Error |
|---|---|---|---|---|
| 3 | 1 | [0.70608] | 0.4 | 0.833 |
| 3 | 2 | [0.53732  0.14595] | 0.4 | |
| 3 | 3 | [0.96676  0.15111  0.0015139] | 0.4 | |
| 10 | 1 | [0.76393] | 0.667 | 0.833 |
| 10 | 2 | [0.2953  0.41039] | 0.667 | |
| 10 | 3 | [0.97014  0.29259  0.080068] | 0.667 | |
| 20 | 1 | [0.028529] | 0.267 | 0.833 |
| 20 | 2 | [0.93581  0.20714] | 0.233 | |
| 20 | 3 | [0.18231  0.09461  0.68031] | 0.233 | |

**Trace**

| α | n-gram | GANED $\lambda_n$ | Classification Error | MINDIST Classification Error |
|---|---|---|---|---|
| 3 | 1 | [0.96149] | 0.27 | 0.54 |
| 3 | 2 | [0.80699  0.14789] | 0.26 | |
| 3 | 3 | [0.89336  0.01668  0.3959] | 0.26 | |
| 10 | 1 | [0.76432] | 0.08 | 0.42 |
| 10 | 2 | [0.42505  0.64252] | 0.09 | |
| 10 | 3 | [0.95513  0.93675  0.99434] | 0.04 | |
| 20 | 1 | [0.92115] | 0.1 | 0.36 |
| 20 | 2 | [0.89948  0.8597] | 0.12 | |
| 20 | 3 | [0.69135  0.21079  0.9382] | 0.12 | |

As we can see from the results, the classification errors of GANED are smaller than those of MINDIST for all the datasets and for all values of the alphabet size. The results of other datasets in the archive were similar.

## 5 Conclusion and Perspectives

In this paper we presented a new normalized edit distance. This new distance, GANED, is related directly to the edit distance and it takes into account the length of the two strings. The particularity of our new distance is that it uses an optimization algorithm; the genetic algorithms, to set the values of its parameters. We tested the new distance by comparing it to another distance applied to strings and we showed how our new distance GANED has a better performance.

The new distance was applied in this work to symbolically represented time series. However, we believe other applications might be more appropriate for our new distance.